\documentclass{jfm}
\usepackage{graphicx}

\graphicspath{ {./images/} }
\usepackage{epstopdf, epsfig}
\usepackage{amsmath}
\usepackage{amssymb}
\usepackage{cmap}
\usepackage{lgrind}
\usepackage{float}

\usepackage[square,comma,numbers]{natbib}

\shorttitle{LR and sparse compression}
\shortauthor{P. Kaloshin}

\title{Convolutional neural networks compression with low rank andsparse tensor decompositions}

\author{Pavel Kaloshin
  \corresp{\email{kaloshinpavel@gmail.com}}}

\affiliation{Master’s Educational Program: Data Science\\Skoltech, Russia}

\begin{document}

\maketitle

\begin{abstract}
Convolutional neural networks show outstanding results in a variety of computer vision tasks.However, a neural network architecture design usually faces a trade-off between model performance and computational/memory complexity. For some real-world applications, it is crucial to develop models, which can be fast and light enough to run on edge systems and mobile devices.However, many modern architectures that demonstrate good performance don’t satisfy inference time and storage limitation requirements. Thus, arises a problem of neural network compression to obtain a smaller and faster model, which is on par with the initial one.

In this work, we consider a neural network compression method based on tensor decompositions. Namely, we propose to approximate the convolutional layer weight with a tensor, which can be represented as a sum of low-rank and sparse components. The motivation for such approximation is based on the assumption that low-rank and sparse terms allow eliminating two different types of redundancy and thus yield a better compression rate. An efficient CPU implementation for the proposed method has been developed. Our algorithm has demonstrated up to 3.5x CPU layer speedup and 11x layer size reduction when compressing Resnet50 architecture for the image classification task.
\end{abstract}

\tableofcontents
\newpage 
\section{Introduction}
\subsection{Convolutional neural networks}
Convolutional neural network is an extremely important class of machine learning models for computer vision area and state of the art architectures for various vision tasks: YOLOv3 \cite{yolo3} for detection, Densenet \cite{densenet} and inception Resnet \cite{inception} for classification, Deeplab \cite{deeplab} for semantic segmentation and so on. The main building block of this type of architectures is convolutional layer - 4D tensor W that (basically) transforms 3D input tensor X into 3D output Y as follows: 
$$Y_{ijt} = X_{i+m-\frac{M}{2},j+n-\frac{N}{2},k}W_{rnkt},$$
where \textit{M} and \textit{N} are spatial dimensions sizes of the convolutional kernel and m, n are running indices in tensor contraction (Einstein notation). This is extremely useful operation that allows to obtain translation-invariant geometrical features. In modern network architectures, convolutional layers are stacked hierarchically to extract multiscale visual patterns then used for the specific task. For example, there are multiple types of Resnet containing 16, 50, 152 convolutional layers. The convolution operation is well-parallelisable thus can be efficiently inferred on GPUs. 
 
\subsection{Network compression motivation}
Computer vision tasks arise in different areas of life: industry wants Software as a Service (SaaS) systems to process their video streams for production quality control and edge-solutions for smart cameras, that react immediately on dangerous situations, IT companies want efficient solutions for processing user videos for adverse content detection, usual people want their smartphones to translate the text on the video from camera immediately, without sending it somewhere. These give two major settings: model runs on GPU in the cloud or on some specific device with limited computational resources. Both settings set limitations on the models: in cloud it is better to run model as fast as possible to process more data per unit time thus reducing the required hardware cost whereas on edge devices with limited RAM, memory and parallel cores the models should be small in size, cache-efficient and have as few operations as possible, while preserving the same quality. To meet these limitations we need efficient and platform-specific ways to compress neural networks.

\subsection{Low rank and sparse decomposition}
Matrix or tensor decomposition can be thought as an approach to decrease redundancy in the representation by projecting it onto some restricted domain - for low rank decomposition, matrix (or tensor) is represented as a linear combination of rank 1 elements of the same shape, and for sparsification, only fixed number of nonzero elements are left. However, this projection leads to information loss, especially when redundancy is complexm e.g. when the original matrix consists of few linearly dependant columns and some outliers. For these complex cases it may be reasonable to represent redundancy in multiple ways simultaneously - for example, decompose the original matrix (or tensor) into a sum of low rank and sparse components. This approach may lead to more compact and precise approximations.
\section{Literature review}
\subsection{Network compression methods}
In early researches it was shown that trained neural networks have huge amount of redundancy: for example authors of \cite{hassibi_second_nodate} managed to successfully remove most of the network weights with almost no quality drop. In future works researchers explore multiple approaches affecting different types of redundancy, which can be often applied together and provide higher level of compression and computational efficiency. The major directions of research in this field are described below. 

\subsubsection{Quantization}
The idea behind this approach is to store weights and process computations with less numerical precision \cite{rounding}, \cite{mixed_precision}, which can directly speed up computations and decrease storage space for different devices simultaneously \cite{quantization_mobile}. However, naive rounding starts to affect the model quality when the number of bits per weight become low. There are multiple approaches to compress the weights further, including hashing \cite{hashnet}, clustering, entropy encoding \cite{quantization_limits}, or trainable quantizaiton \cite{trainable_quantization}. The extreme case of quantization is binarization - restricting the network weights to be 0 or 1 \cite{binarization}, \cite{bulat_xnor-net_2019}. This approach leads to extremely small models, but the quality drop is quite high for now.
\subsubsection{Knowledge distillation}
It is hard to train deep networks efficiently and there historically are some tricks to withstand it \cite{residual}, \cite{batchnorm}, \cite{googlenet}. However, well-performing network still might be deeper than required, so we may replace it's parts with smaller ones trained to reproduce the original parts output \cite{knowledge_distillation}.

\subsubsection{Pruning and sparsification}
In \cite{brain_damage} the authors removed part of network weights using second order derivatives. Other approaches include magnitude pruning \cite{sp_state}, variational dropout \cite{molchanov_variational_2017}, learnable sparsity \cite{liu_learning_2017}, \cite{lee_differentiable_2019}, fisher information based pruning \cite{theis_faster_2018}. Worth mentioning that sparse layers are typically hard to be implemented efficiently on devices in terms of inference speed, so there also is a branch of research about structured sparsity \cite{wen_learning_2016}, \cite{luo_thinet_2017}.
There also is a specific topic on neural network sparsity related to "lottery ticket hypothesis" \cite{lottery_ticket}, \cite{ramanujan_whats_2019}. We can obtain the sparse mask for trained network weights that have almost the same quality as the original network, moreover, we can apply this mask to the original weight initialisation and obtain nonzero quality without any training. This gives us a clue about the existence of sub-network in randomly initialised network, that can already solve the required task efficiently. \cite{NIPS2019_8620} investigated these masked weights behaviour even further and give some insights about network training, for example, that weights often do not change their sign during optimization. In \cite{malach_proving_2020} lottery ticked hypothesis was proved theoretically along with the fact that it is not satisfied for structurally sparse mask.

\subsubsection{Tensor decompositions}
Weights of a network can be presented as 2D tensors (for fully connected layers) or 4D tensors (for convolutional layers). These tensors might be approximated with low rank representations, that have less parameters to store. It can be done with several well-known tensor decompositions \cite{tai_convolutional_2016}, \cite{lebedev_speeding-up_2015}, \cite{sepfilters}, \cite{svd_decomp}. Authors in \cite{kossaifi_t-net_2019} represented all the network weights using single tensor and showed that the compression of this tensor is more effective than the one by one layer compression.

\subsection{Speedup limitations}
FLOPS is a prefered metric to access the compressed model inference speed. However, sometimes models with a huge number of FLOPS can be even faster, because their structure is optimized for computations \cite{ma_shufflenet_2018}. That is why network compression method should provide parameter reduction along with efficient inference schema in terms of model speedup.

\subsection{Low rank and sparse decompositions}
Sparse term in low rank matrix decomposition naturally appeared to make PCA more resistant to outliers or missing data:
\begin{subequations}
\begin{alignat}{2}
&\min_{L, S}        &\qquad& rank(L) + \lambda \|S\|_0 \label{eq0:optProb}\\
&\text{subject to} &      & W = L + S,\label{eq0:constraint1}
\end{alignat}
\end{subequations}

This is a well known optimization task for Robast PCA \cite{RPCA}. There is a brunch of studies related to complexity \cite{RPCA_complexity}, modifications \cite{rpca-noise} and applications, such as face completion \cite{rpca-faces}, background subtraction \cite{rpca_bgsub}, anomaly detection \cite{RPCA_anomaly}, etc.

However, these matrix methods are not straightforward to apply for four-dimensional tensors, which is crucial for compression of convolutional layers kernels. There are some adaptations (\cite{tensorRPCA} for example), that are mostly based on well known low rank tensor decompositions and iterative optimization algorithms like alternating least squares. To the best of my knowledge, there is no information about applying these methods to neural networks compression.
\section{Problem statement}
This work focuses on convolutional neural network compression via low rank and sparse tensor decomposition. This chapter will introduce notation and overall procedure. 

\subsection{Neural network compression} \label{compression_process}
Let X denote some set of samples and Y - the true values on these samples. Convolutional neural network is a function $F(W_1, ... W_n, X) = \hat{Y} $ that is used to approximate Y (n is number of convolutional layers, $W_i$ is an order 4 tensor representing weights of i-th convolutional layer). The quality of approximation is measured with quality function Q, that is to be maximized during training:
$$W_1^*, ..., W_n^* = \text{argmax}_{W_1, ... W_n}\ Q(Y, F(W_1, ..., W_n, X))$$
Let $P(W_i) \rightarrow \mathbb{N}$ be a function that returns the number of parameters in $W_i$. The task is to find representations $\hat W_i$ with less number of parameters than that of the original layers, but preserve the quality of the model up to some pre-defined bounds. If the number of non-convolutional parameters in model is M, the overall optimization task is formulated as following:

\begin{subequations}
\begin{alignat}{2}
&\max_{\hat W_1, ..., \hat W_n}        &\qquad& \frac{\sum_{i=1}^n P(W_i^*) + M}{\sum_{i=1}^n P(\hat W_i) + M}\label{eq:optProb}\\
&\text{subject to} &      & Q(F(\hat W_1, ..., \hat W_n, X), Y) > Q(Y, Y^*) - s,\label{eq:constraint1}
\end{alignat}
\end{subequations}
\noindent
where $Y^* = F(W_1^*, ..., W_n^*, X)$. However, in practice we cannot solve this task with respect to all the network parameters simultaneously. Instead we iteratively optimize over single layers:
\begin{subequations}
\begin{alignat}{2}
&\max_{\hat W_i}        &\qquad& \frac{\sum_{i=1}^n P(W_i^*) + M}{... + \hat W_i + ... + M}\label{eq2:optProb}\\
&\text{subject to} &      & Q(F(..., \hat W_i, ..., X), Y) > Q(Y, Y^*) - s_i,\label{eq2:constraint1}\\
& &      & P(\hat W_i) < P(W_i^*), \label{eq2:constraint2}
\end{alignat}
\end{subequations}

To use this schema one needs to choose the sequences $\{i_j\}_{j=1}^N$ (order of layers) and $s_{i_j}$ (allowed accuracy drop on each step).

\subsection{Tensor decomposition}
W is a order 4 tensor to be approximated with sum of a low rank component L and a sparse component S, that have as few parameters as possible. The optimization task looks as following: 
\begin{subequations}
\begin{alignat}{2}
&\min_{L, S}        &\qquad& \|W - L - S\|\label{eq3:optProb}\\
&\text{subject to} &      & \text{rank}(L) = r,\label{eq3:constraint1}\\
& &      & \text{card}(S) = c,\label{eq3:constraint2}
\end{alignat}
\end{subequations}

\noindent
Here $\text{card}(S)$ is the fraction of nonzero elements in S, $\text{rank}(L)$ is a rank of the 4D tensor L. A rank of a tensor can be defined in multiple ways. In this work the definition coming from CP decomposition is used:
$$L_{ijkt} = A_{ir}B_{jr}C_{kr}D_{tr},$$
where A, B, C, D are matrices and r is a rank of the tensor L.
Let I, J, K, T denote the original tensor dimensions sizes. Then the original number of parameters of W will be $P(W) = IJKT$, and for the decomposed terms we have $P(L) = r(I + J + K + T)$, $P(S) = \alpha IJKT$, where $\alpha$ represents the fact that we need to store the value of sparse tensor along with it's index.

There exist different approaches for CP decomposition (for example, \cite{cpdec1}, \cite{cpdec2}) and, to the best of my knowledge, all of them are iterative. Any iterative approach can be easily extended to low rank and sparse version. On each step i:
\begin{enumerate}
    \item $L_i = update(L_{i-1}, W - S_{i-1})$
    \item $S_i = P_c(W - L_i)$
\end{enumerate}

Here $P_c(A)$ is projection of some matrix A on the space of matrices with a fixed cardinality c. The nonzero elements a choised by the maximum absolute value.

\subsection{CP-decomposed convolution} \label{cp_conv}

\begin{figure}
        \centering
        \includegraphics[width=\textwidth]{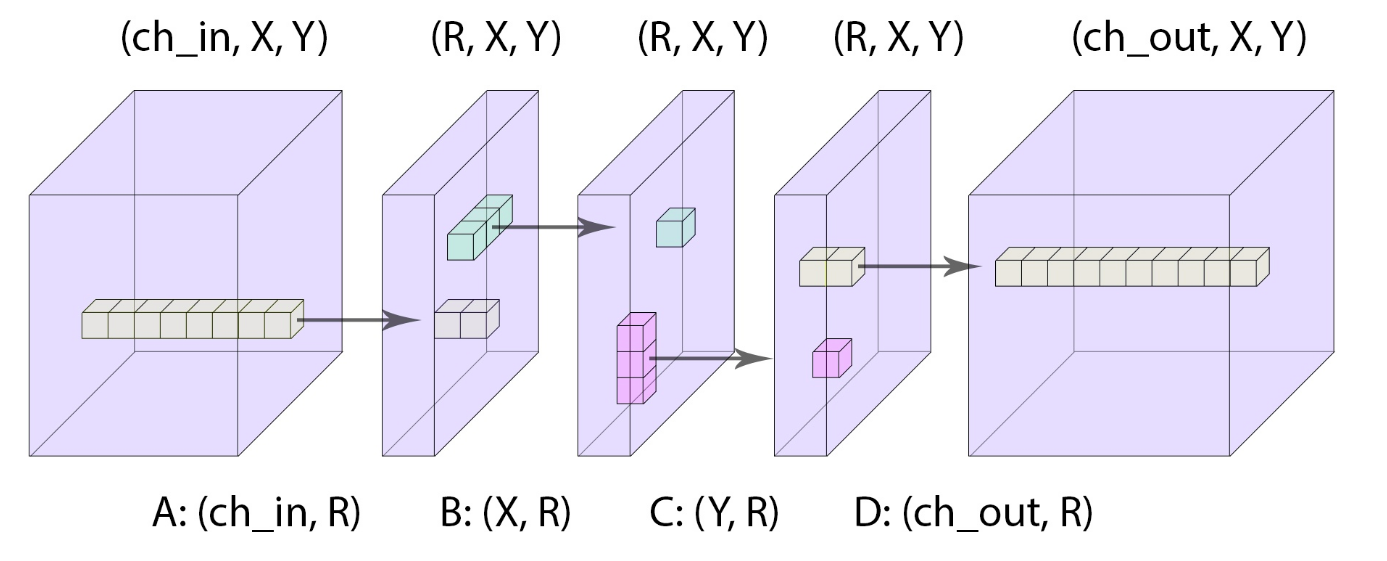}
        \caption{CP convolution}
        \label{fig:cp-conv}
\end{figure}

We have the low-rank part of the convolution in the following form:.
$$L_{ijkp} = A_{ir}B_{kr}C_{pr}D_{jr}$$
This type of decomposition allows efficient implementation of convolution with L. We consider original convolution weight tensor $W_{ijkp}$ applied to the input tensor $X_{smt}$. The first two indices of X (s and m) denote spatial dimensions x and y, and the 3rd index "t" is for number of channels. As for W, indices i and j are for input and output number of channels and indexes k and p are for spatial kernel dimensions (kernel size is $N_k$x$N_p$ ). The result of convolution is (we assume that k=0,p=0 is center of convolutional kernel to make the notation easier):
$$Y_{smj} = X_{(s+k)(m+p)i}\ W_{ijkp}$$
We hence can apply the decomposition terms of L sequentially, as four consequent convolutions (fig. \ref{fig:cp-conv}).
$$Y^A_{smr} = X_{smi}\ A_{ir}$$
This is equivalent to convolution with kernel of shape (i,r,1,1) (also known as 1x1 convolution). 
Next two matrices affect spatial components:
$$Y^B_{smr} = Y^A_{(s+k)mr}\ C_{kr}$$
$$Y^C_{smr} = Y^B_{s(m+p)r}\ C_{pr}$$
The corresponding convolutional kernel shapes are (r,r,k,1) and (r,r,1,p). And the last multiplier is 1x1 convolution with kernel shape (r,j,1,1):
$$Y^A_{smj} = Y^C_{smr}\ D_{jr}$$

\subsection{Efficient sparse convolution} \label{sparse_conv_algorythm}
\begin{figure}
\centering
  \begin{minipage}[b]{0.35\textwidth}
    \includegraphics[width=250pt]{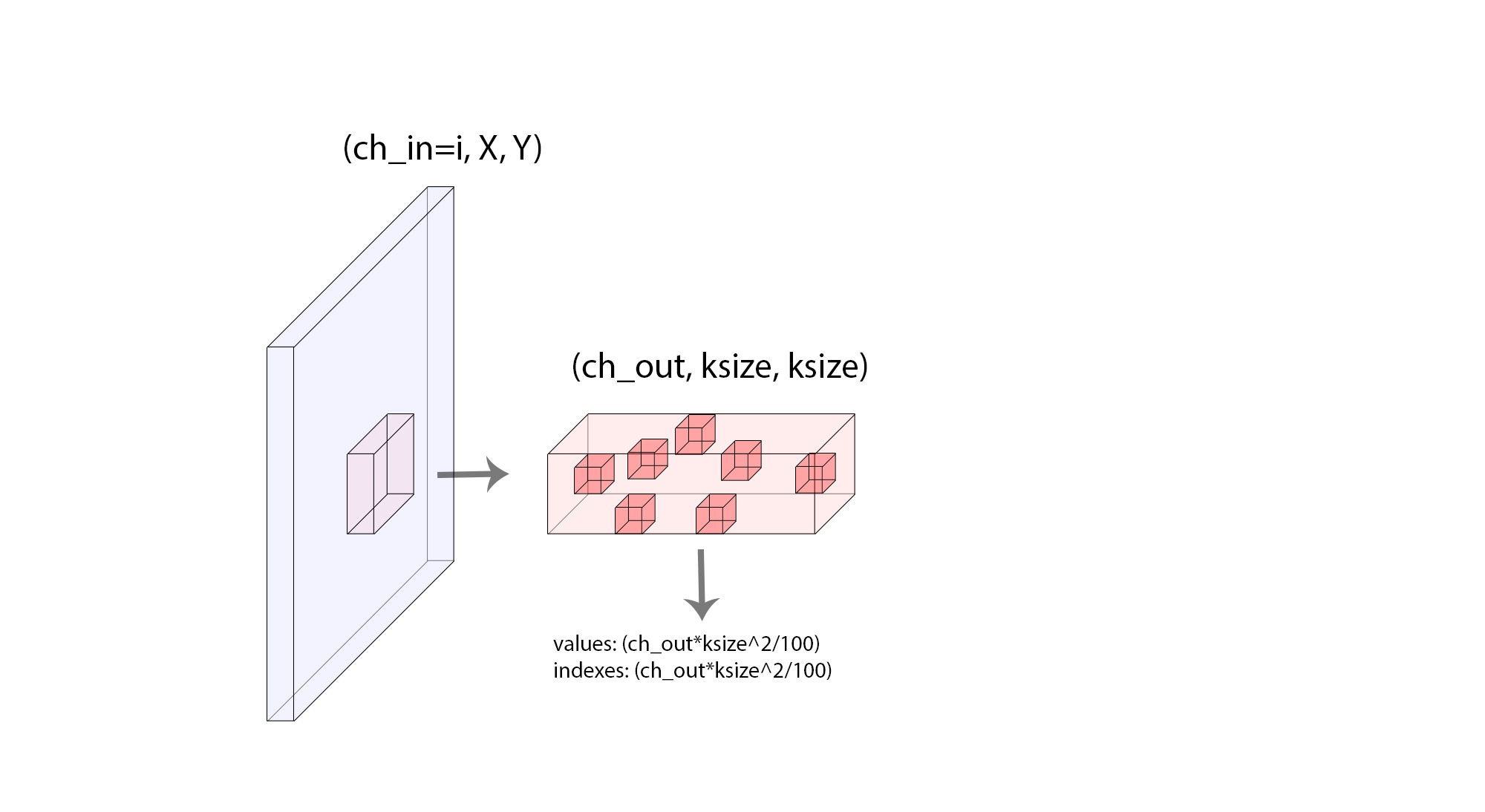}
    \caption{Sparse slice}
    \label{fig:sp-slice}
  \end{minipage}
  \hfill
  \begin{minipage}[b]{0.35\textwidth}
    \includegraphics[width=250pt]{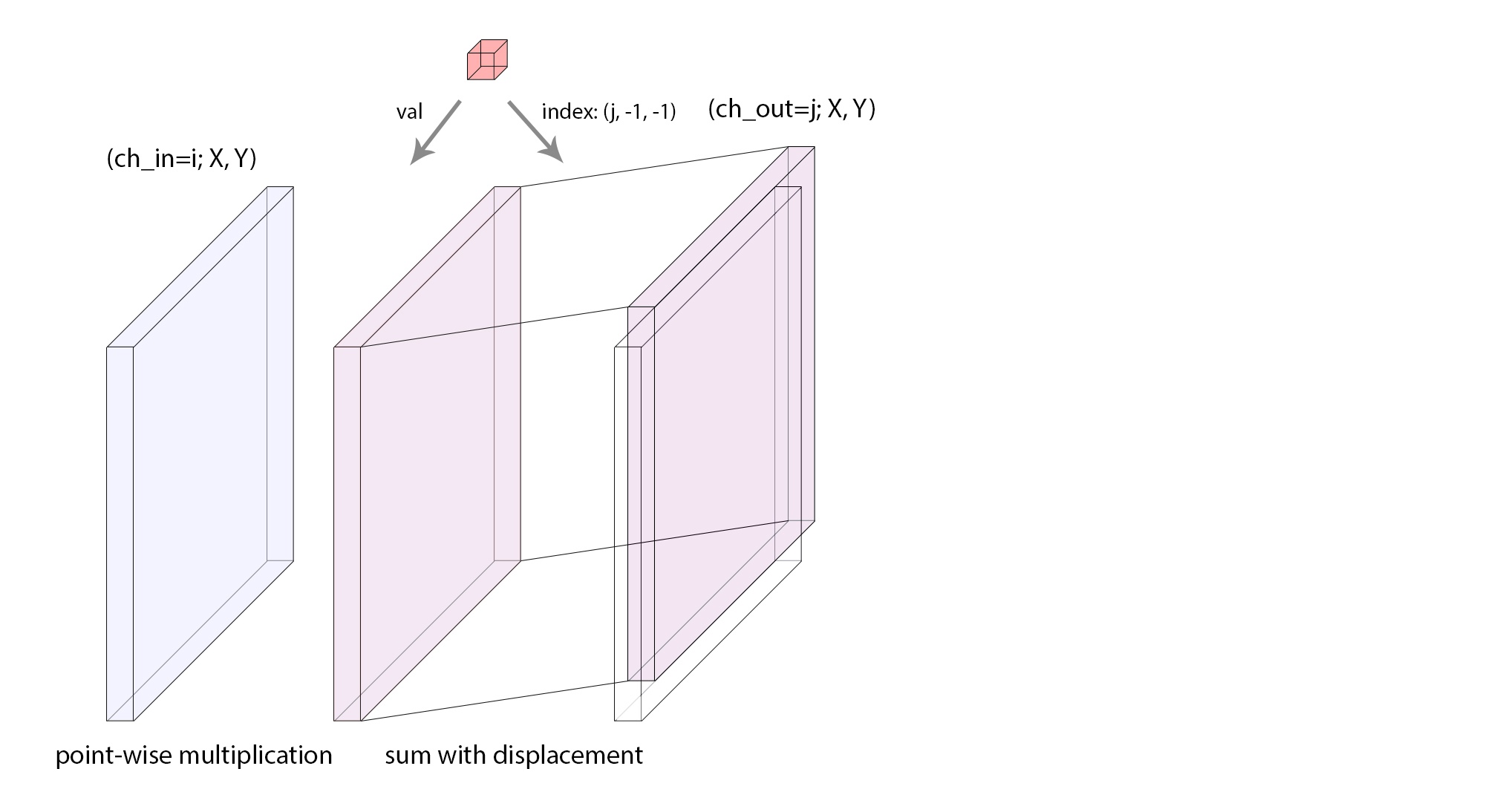}
    \caption{Nonzero element}
    \label{fig:sp-nonzero}
  \end{minipage}
\end{figure}

Sparse convolutions require much less computations, than the dense ones, because only nonzero weights are required to be processed. However, it is difficult to gain from this in terms of inference time if the sparse pattern is random due to data reading issues. It is much faster to read data from memory consequently than from unordered places that nonzero elements point to. However, if the matrix is very sparse, it is possible to develop a cache-friendly way to perform this convolutions and gain some speedup. 

Let's consider i-th input channel with the corresponding i-th slice of sparse kernel (Fig. \ref{fig:sp-slice}). For convolution operation each value in input channel should be multiplied by each value in this kernel slice and then stored properly. To do this, we store this slice as two arrays, one with values of nonzero elements and one with output channel number and position in kernel, both can be stored in one int16 number. Then we iterate through these arrays, multiply each element in the input by value and add the result to the corresponding channel of output with the spatial displacement related to the position of the element in kernel. For example, if the position of the element in the kernel is (1, 1), the result of the multiplication should be displaced by -1, -1, as shown on Fig. \ref{fig:sp-nonzero}.

This schema allows consequent reading of the input data, which is much bigger than the sparse kernel, thus making the sparse convolution operation cache-friendly.

\subsection{Fine-tuning procedure}
The optimization task \ref{eq:optProb} allows to obtain $\hat W = L + S$ approximation of W, which is good in terms of L2 norm of residual thus the norm of difference between outputs is not that big, but this is not directly related to the overall model quality, thus leading to the performance drop. To mitigate this the decomposed layer should be fine-tuned using ordinary back-propagation with respect to the original loss. During the fine-tuning procedure the original model weights (weights in the layers that are not decomposed yer) should be frozen to prevent the loss of intermediate representations distribution details learned by the original model.
\section{Experimental setup}
\subsection{Network architecture: resnet50}

Resnet50 was chosen for experiments as widely used convolutional network architecture. It contains 52 convolutional layers and one fully connected layer, which is not compressed in this work, resulting in 26 128 695 trainable parameters in total, 23 454 912 of them are directly related to convolutional kernels. We use implementation from pytorch \cite{pytorch} pretrained on Imagenet dataset \cite{imagenet}, a large classification dataset containing 1 281 167 train and 50 000 val images divided into 1000 classes to 76.13\% top1 accuracy and 92.862\% top5 accuracy.

\begin{figure}
        \centering
        \includegraphics[width=300pt]{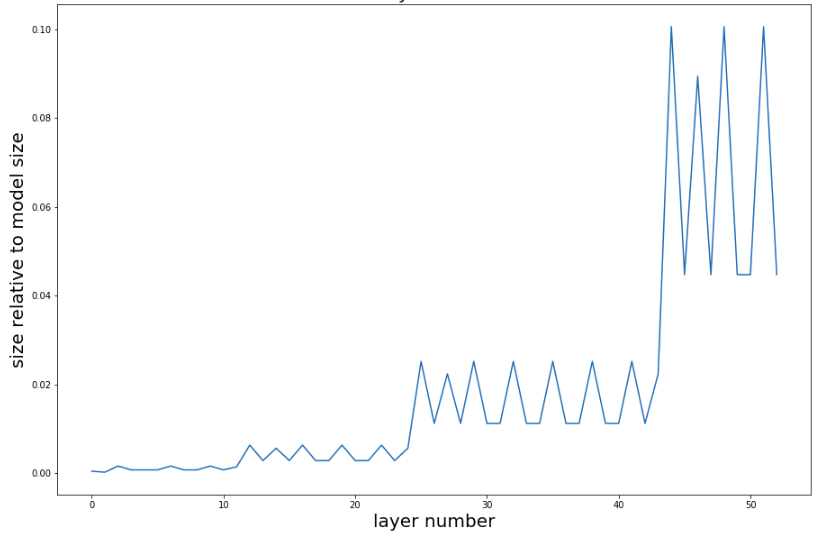}
        \caption{Resnet50 layers sizes}
        \label{fig:lsizes}
\end{figure}

\begin{figure}
    \centering
  \begin{minipage}[b]{200pt}
    \includegraphics[width=200pt]{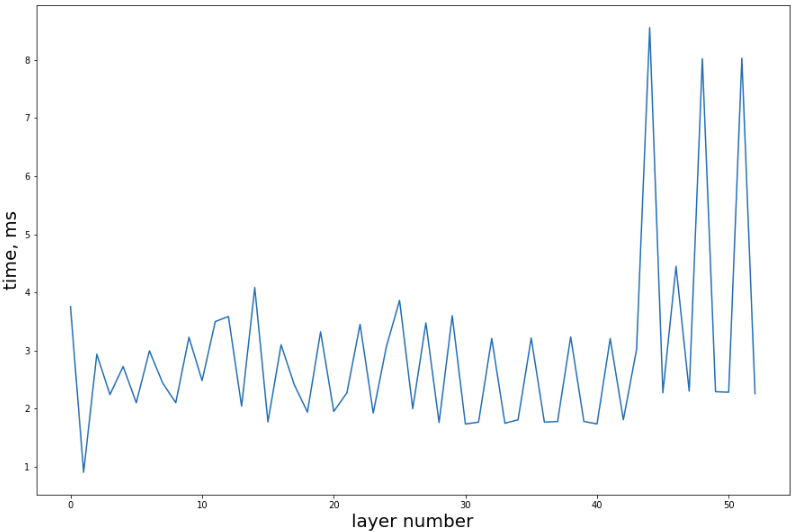}
    \caption{Resnet50 layers CPU inference times}
    \label{fig:cpu_time}
  \end{minipage}
  \hfill
  \begin{minipage}[b]{200pt}
    \includegraphics[width=200pt]{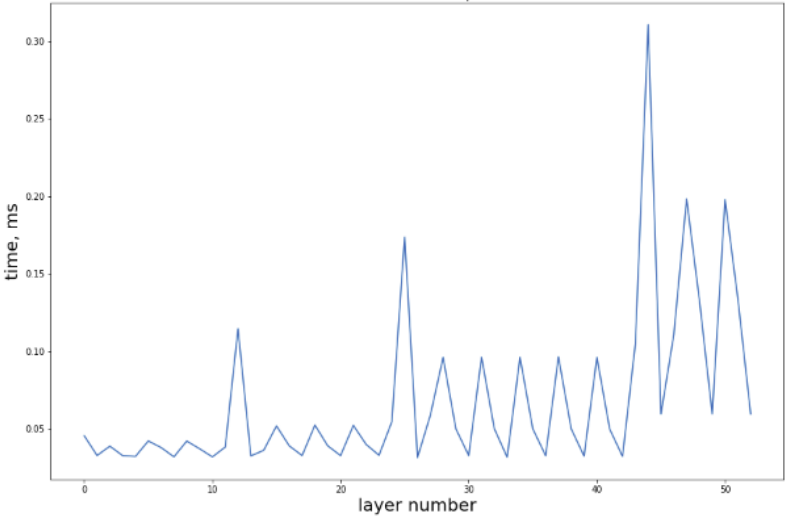}
    \caption{Resnet50 layers GPU inference times}
    \label{fig:gpu_time}
  \end{minipage}
\end{figure}

The network has 53 convolutional layers (Table \ref{table:lshapes}) with varying size and inference speed, which are generally increasing to the end of the network (Figures \ref{fig:lsizes}, \ref{fig:cpu_time}, \ref{fig:gpu_time}) therefore it is better to allow more performance drop for the last layers (thus compressing them better). 

\subsection{Tensor decomposition reparameterization} \label{reparametrization}
We seek the decomposition in Prob. \ref{eq2:optProb} that gives maximum compression, but still reasonably preserves the model quality. Unfortunately, problem \ref{eq3:optProb} is parametrized with rank r and cardinality c, that are not informative in terms of the model quality. Instead it may be better to fix the relative residual norm and search the best compression possible:


Thus we obtain the new parameter $\epsilon$, which is better correlated with model performance and makes the decomposition selection easier. \textbf{The cardinality value c is fixed to 1\% for all the experiments}.

\subsection{CP components refinement}

CP decomposition can be represented as a sum of rank 1 tensors:
$$L_{ijkt} = \sum_{r=1}^R A_{ir}B_{jr}C_{kr}D{tr} \Rightarrow L = \sum_{r=1}^R a_r \otimes b_r \otimes c_r \otimes d_r,$$
where $a_r,\ b_r,\ c_r,\ d_r$ denote r-th columns of A, B, C, D and $\otimes$ is Kronecker product. Large norms of these rank 1 tensors might lead to numerical instability of the approximation, thus it can be harder to fine-tune. Authors of \cite{EPC} developed an algorithm that minimizes the sum of their norms $\sum_{r=1}^R \|a_r \otimes b_r \otimes c_r \otimes d_r\|_F^2$ while preserving the decomposition error. This improvement leads to better model quality after the decomposed layer fine-tuning (Fig. \ref{fig:epc}).
\section{Experiments and results}
\subsection{Fixed error decomposition}
The first idea is to keep the decomposition error bound $\epsilon$ fixed to curtain values and obtain the baseline compression rates. For each value of $\epsilon$, the layers were compressed sequentially from 0th to 52nd. The fine-tuning was performed with Adam optimizer and one cycle learning rate schedule \cite{one-cycle} from fastai deep learning framework, 2 epochs per decomposed layer. During each layer fine-tuning, all the previously decomposed layers were also updating to prevent error accumulation, while the original model layers were frozen to preserve the original layers output distibutions. The results summarized in table \ref{table:fix_decomp}  are not very good  due to huge accuracy drop, but give us a clue about the fact that at least some layers can mitigate quite high values of decomposition error and still preserve the output distribution reasonably in terms of overall model performance.

\subsection{Speedup bound} \label{bound}
Before the whole network compression it is reasonable to estimate the level of speedup which is possible to obtain on different levels of compression. In the following subsections the low rank and sparse parts speedups are investigated separately. 
\subsubsection{Low rank component}
\begin{figure}
        \centering
        \includegraphics[width=0.7\textwidth]{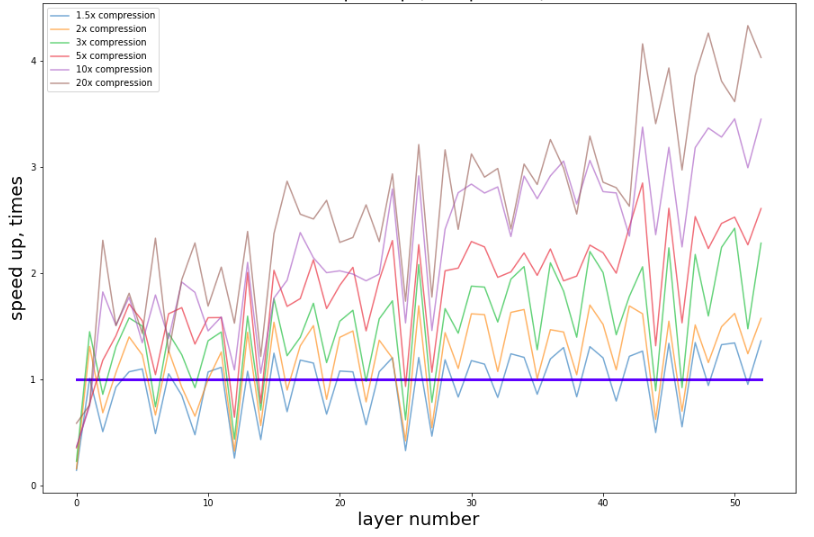}
        \caption{Resnet50 LR layers speedup on Intel Core i3-8130U CPU} 
        \label{fig:lr_speedup_x1}
\end{figure}

\begin{table}
\centering
\begin{tabular}{ |c|c| } 
 \hline
 compression & speedup \\
 \hline\hline
1.5x & 1.08x \\
2x &  1.2x \\
3x & 1.39x \\
5x & 1.66x \\
10x & 2.03x \\
20x & 2.33x \\
 \hline
\end{tabular}
\caption{Decomposition with fixed error bound}
\label{table:lr_speedup_x1}
\end{table}

For each layer CPU inference time was measured. Assuming that the compression rate value is fixed, each layer of the network was represented in CP form with rank corresponding to that compression rate, as was described in Section \ref{cp_conv}, and inference time of that decomposed layers was measured. For fixed compression rate the low rank inference time divided by original layer inference time is plotted on Fig. \ref{fig:lr_speedup_x1}. The whole network compression-speedup relation is presented in Table \ref{fig:lr_speedup_x1}. Also the same measurements were taken for the x2 and x3 input size, Fig. \ref{fig:lr_speedup_x2}, Table \ref{table:lr_speedup_x2} and Fig. \ref{fig:lr_speedup_x3}, Table \ref{table:lr_speedup_x3} correspondingly. All the convolutions were performed using pytorch deep learning framework.

\subsubsection{Sparse component}

\begin{figure}
    \centering
  \begin{minipage}[b]{0.4\textwidth}
    \includegraphics[width=180pt]{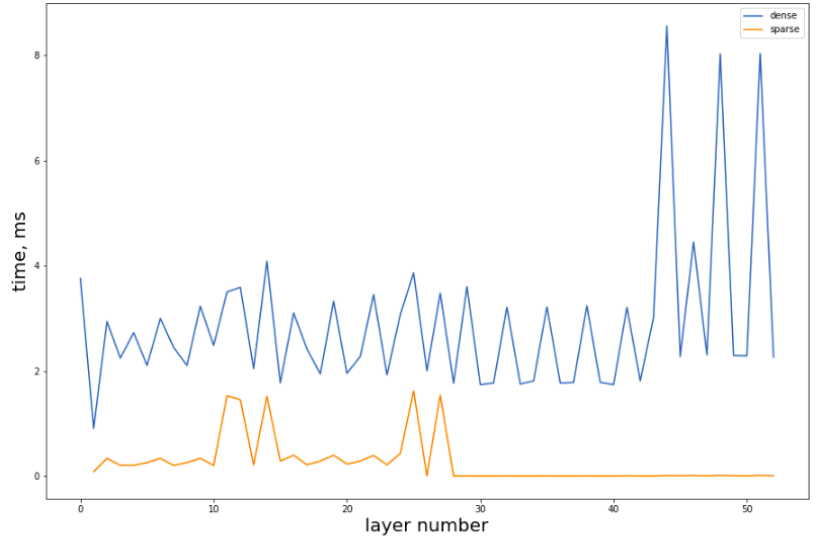}
    \caption{Sparse convolution inference time on Intel Core i3-8130U CPU}
    \label{fig:sp-cpu-time}
  \end{minipage}
  \hfill
  \begin{minipage}[b]{0.4\textwidth}
    \includegraphics[width=180pt]{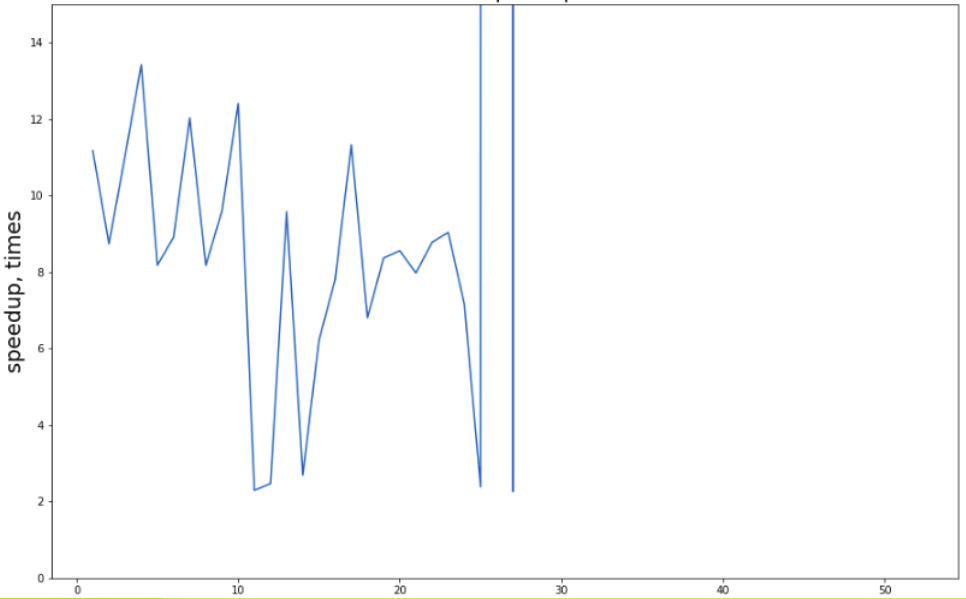}
    \caption{Sparse convolution speedup on Intel Core i3-8130U CPU (compared to dense convolution of the same shape)}
    \label{fig:sp-cpu-speedup}
  \end{minipage}
\end{figure}

\begin{figure}
        \centering
        \includegraphics[width=0.5\textwidth]{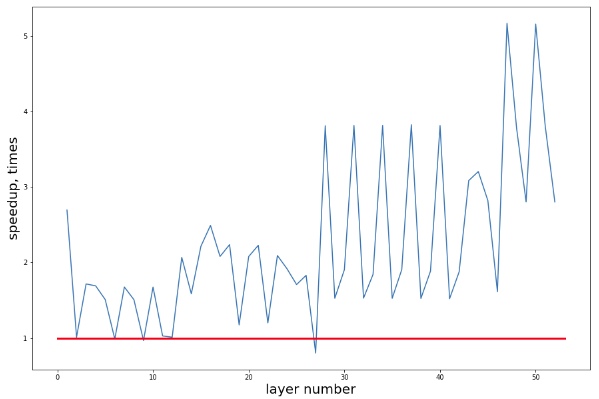}
        \caption{Sparse convolution on Intel Core i3-8130U CPU compared to dense convolution on Nvidia 1080Ti GPU speedup}
        \label{fig:gpu_speedup}
\end{figure}

The convolution algorithm described in section \ref{sparse_conv_algorythm} was implemented on C with the same number of nonzero elements in input channel-wise slices for simplicity. Each layer of the Resnet50 was sparsified leaving 1\% of nonzero elements and its inference time was compared to the original layer inference time on CPU (Fig. \ref{fig:sp-cpu-time}). The ratio of the sparse layers inference time to the dense layers inference time is plotted on Fig. \ref{fig:sp-cpu-speedup}. The comparison to GPU inference time is plotted on Fig. \ref{fig:sp-gpu-time}.

This algorithm was also implemented for CUDA. GPU speedup is plotted on Fig. \ref{fig:gpu_speedup}.

\subsection{Compression and decomposition error dependency} \label{compression_error_dependecy}
\begin{figure}
        \centering
        \includegraphics[width=0.5\textwidth]{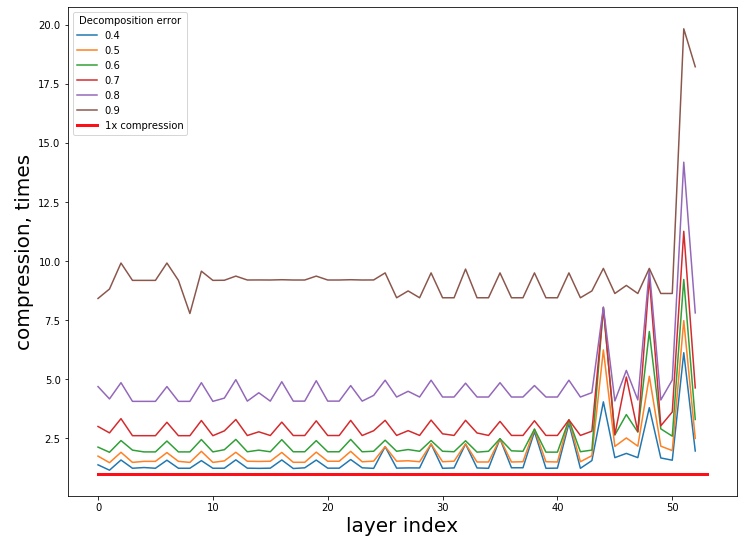}
        \caption{Compression on error dependency}
        \label{fig:err_comp}
\end{figure}
In Section \ref{reparametrization} the new parameter for layer compression was introduced. It is reasonable to investigate its connection to the layer compression level which is the most obvious parametrization for network compression task. For each layer of the Resnet50 the compression was measured for different values of decomposition error. The results are presented on Fig. \ref{fig:err_comp}.

The decomposition error and the compression rate are in close relation for most of the network layers. Nevertheless, fluctuations point to the hypothesis that some layers can be compressed better than the others with the same information loss Also, the last layers reach much higher compression rates with the same decomposition errors, which may mean larger amount of redundancy in them.

\subsection{Decomposition error tolerance} \label{error_tolerance}
\begin{figure}
    \centering
  \begin{minipage}[b]{0.4\textwidth}
    \includegraphics[width=180pt]{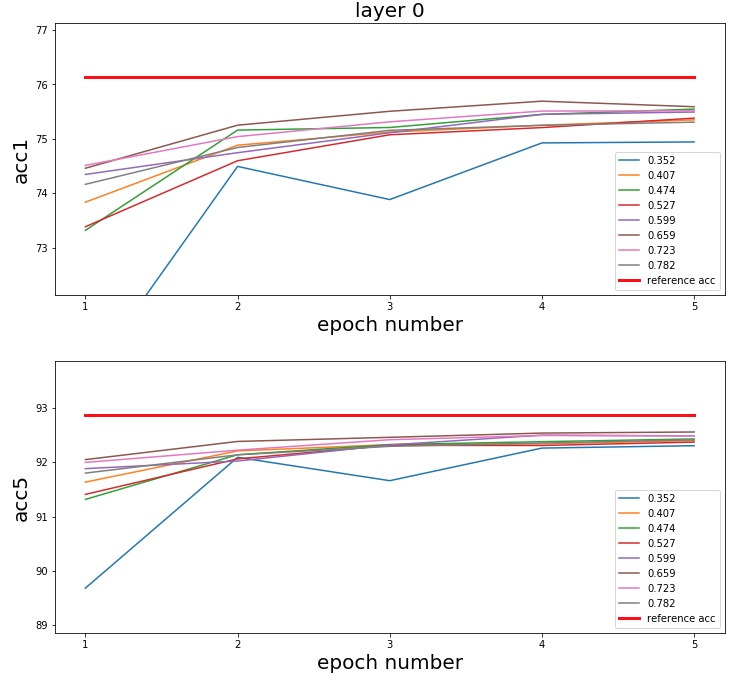}
    \caption{Layer 0 fine-tuning, hard to compress.}
    \label{fig:layer0}
  \end{minipage}
  \hfill
  \begin{minipage}[b]{0.4\textwidth}
    \includegraphics[width=180pt]{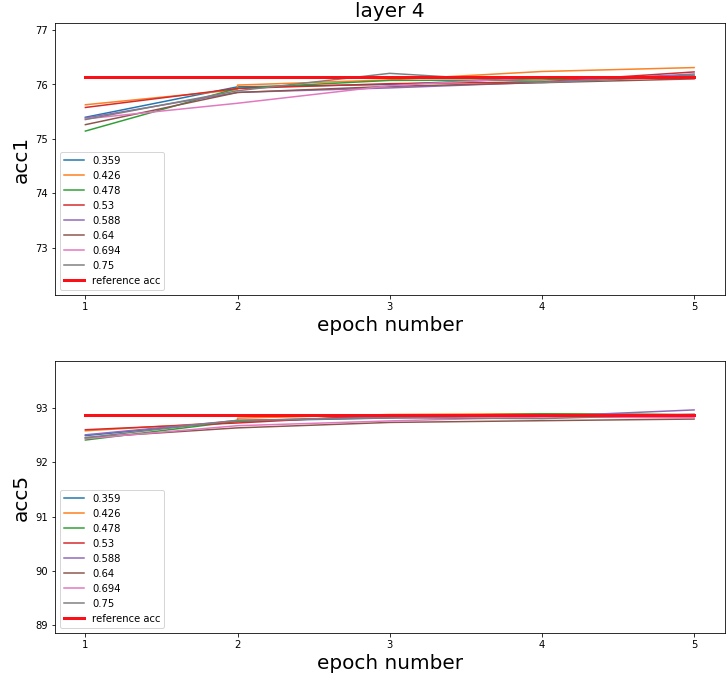}
    \caption{Layer 4 fine-tuning, large layer output redundancy.}
    \label{fig:layer4}
  \end{minipage}
\end{figure}

\begin{figure}
    \centering
  \begin{minipage}[b]{0.4\textwidth}
    \includegraphics[width=180pt]{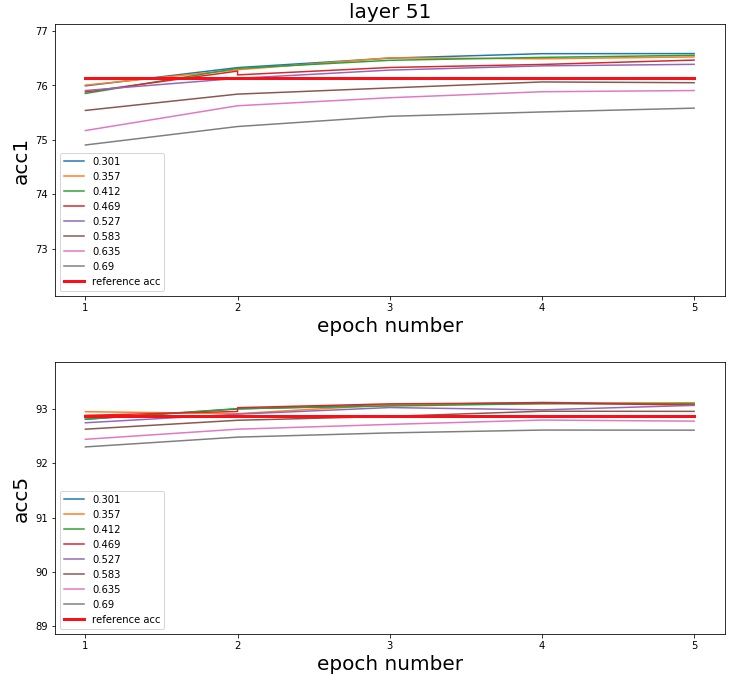}
    \caption{Layer 51 fine-tuning, sufficient redundancy for compression.}
    \label{fig:layer51}
  \end{minipage}
  \hfill
    \begin{minipage}[b]{0.4\textwidth}
    \includegraphics[width=180pt]{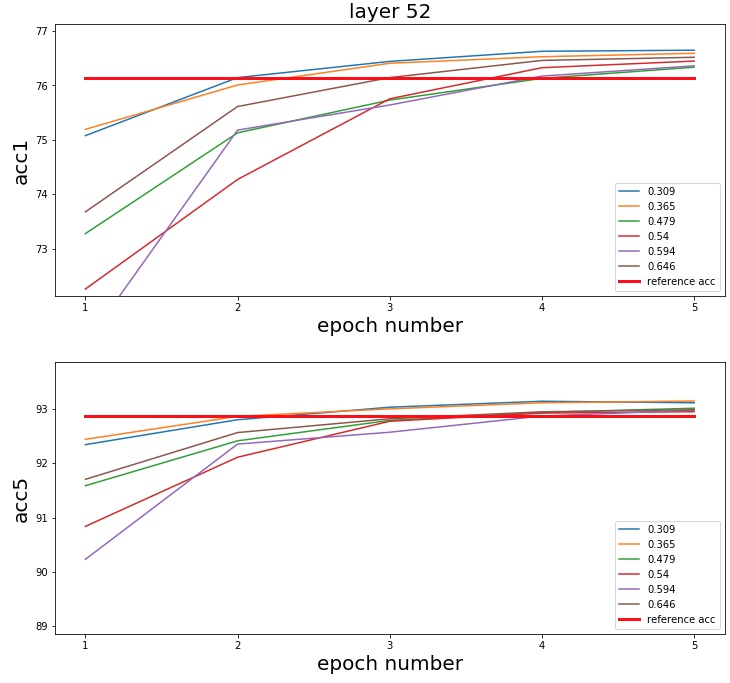}
    \caption{Layer 52 fine-tuning, large layer redundancy, easy to compress.}
    \label{fig:layer52}
  \end{minipage}
\end{figure}

For some layers of Resnet50 few decompositions with different errors were fine-tuned for 5 epochs, each layer was fine-tuned while the other layers from network were frozen to their original weights.The results (Figures \ref{fig:layer0} - \ref{fig:layer52}) show that different layers have different potential for decomposition: layer 0 (Fig. \ref{fig:layer0}) does not tolerate the decomposition at all, layer 4 (Fig. \ref{fig:layer4}) can withstand huge values of decomposition error without much change in model performance (which may be a sign of the fact that large subspace of layer 4 output is irrelevant to model answers), layer 52 (Fig. \ref{fig:layer52}) has a huge redundancy, can be compressed well and after fine-tuning even shows accuracy better than the original layer.

\subsection{Resnet50 compression}
\begin{figure}
    \centering
  \begin{minipage}[b]{0.42\textwidth}
    \includegraphics[width=180pt]{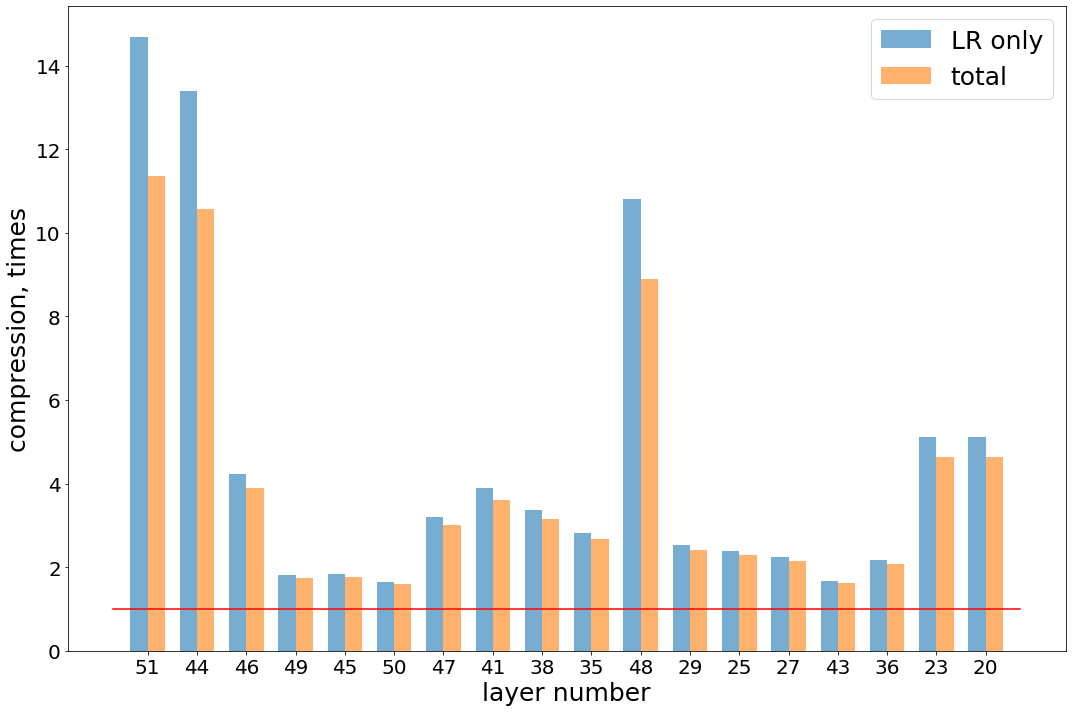}
    \caption{Partial Resnet50 compression}
    \label{fig:rez-comression}
  \end{minipage}
  \hfill
  \begin{minipage}[b]{0.42\textwidth}
    \includegraphics[width=180pt]{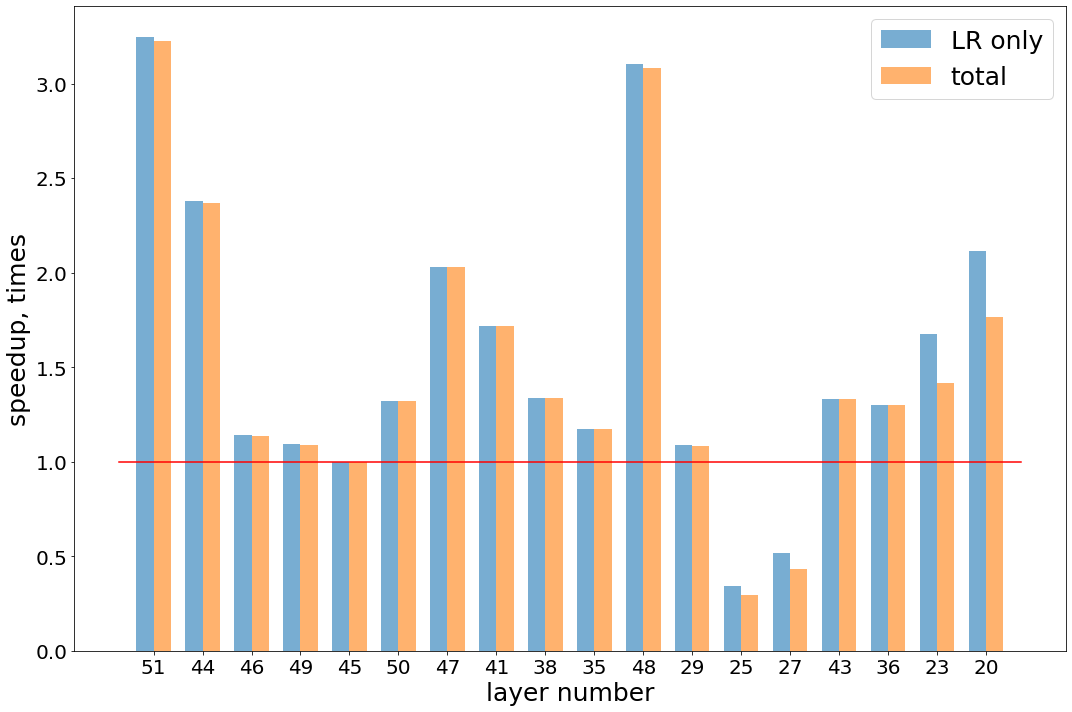}
    \caption{Partial Resnet50 speedup on Intel Core i3-8130U CPU}
    \label{fig:rez-speedup}
  \end{minipage}
\end{figure}

\begin{figure}
        \centering
        \includegraphics[width=0.7\textwidth]{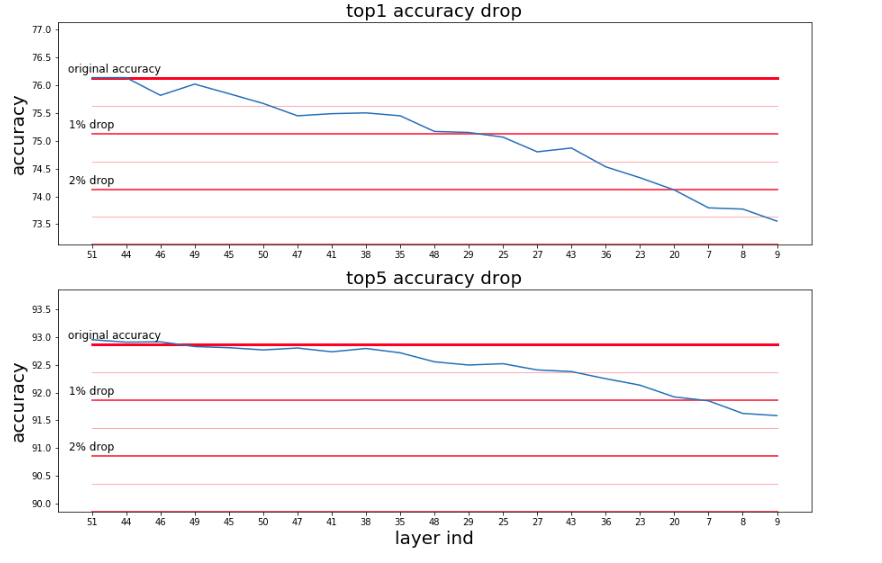}
        \caption{Accuracy drop}
        \label{fig:acc_drop}
\end{figure}

\begin{table}
\centering
\begin{tabular}{ |p{1.8cm}|p{1.8cm}|p{2cm}|p{2cm}|p{2cm}|p{2cm}| } 
 \hline
 top1 \newline accuracy drop & top5 \newline accuracy drop & partial speedup & total speedup & partial \newline compression & total \newline compression \\
 \hline\hline
0.62\% & 0.06\% & 1.53 & 1.08 & 3.47 & 1.59 \\
0.96\% & 0.3\% & 1.35 & 1.11 & 3.78 & 1.9 \\
2.01\% & 0.93\% & 1.18 & 1.09 & 3.4 & 2.15 \\
2.58\% & 1.27\% & 1.21 & 1.12 & 3.4 & 2.16 \\

 \hline
\end{tabular}
\caption{Compression results for different accuracy drops: partial compression (speedup) denotes the compression (speedup) of the compressed part of the network, total compression (speedup) denotes the compression (speedup) of the whole network with this part decomposed.}
\label{table:results}
\end{table}

Top 21 layers of Resnet50 were decomposed using the approach described in Section \ref{compression_process} (layers were being decomposed in size-decreasing order). The fine-tuning for each layer was performed during 5 epochs using SGD with momentum (fixed to 0.9) and learning rate starting from 0.001 and decreasing by half on each epoch. The particular decomposition for each layer was searched by fine-tuning multiple decompositions and choosing the best in terms of compression that preserves the desired model performance. During one layer fine-tuning all the other layers were frozen, it turned out that it was not that important to retrain previously decomposed and fine-tuned layers in terms of model performance, but it takes much more time. Mixed precision technique was used to speed-up the computations \cite{mixed_precision}. The results are summarized on Figures \ref{fig:rez-comression} and \ref{fig:rez-speedup}, blue bars for low rank part only and orange bars for the whole layer decomposition. Cumulative accuracy drop is plotted on Fig. \ref{fig:acc_drop}. The dependency between accuracy drop and compression is summarized in table \ref{table:results}. 

Top9 biggest layers of resnet50 contain 60\% of weights. This part of the network was compressed \textbf{3.47} times giving \textbf{1.5} speedup at the price of \textbf{0.62\%} top1 and \textbf{0.06\%} top5 accuracy drop. The whole network is compressed \textbf{1.59} times giving \textbf{1.08} times speedup.

However, as it was represented in section \ref{bound}, smaller layers are harder to accelerate even when the compression level is reasonable. It can be seen in table \ref{table:results}, that for \textbf{2\%} accuracy drop we have smaller network (\textbf{2.15} times compression), but it is not faster than the network with 9 biggest layers compressed.   
\section{Discussion} \label{discussion}
Different layers tolerate compression differently: some allow multiple times compression rates without accuracy drop, the other cannot be fine-tuned to reasonable performance even with small residual norms. Therefore it is difficult to predict the result of algorithm with given hyperparameters beforehand. However, the size of the layers is increasing rapidly to the end - top 9 biggest layers contain 60\% of the whole model weights, so one may compress only these layers to obtain reasonably small model. Unfortunately, this is not the same for speedup. On the first layers the input is quite large, therefore the inference time for them is not that different to the last layers inference time and they also should be decomposed to obtain reasonable speedup.  
\par
Low sparsity rate (1\% in this work) turns out to be enough to maintain the representation power of decomposed layers and make the decomposition process easier. Moreover, this high sparsity allows efficient implementation on CPU (Fig. \ref{fig:cpu_time}). This implementation is up to 13 times faster for the first half of the resnet50 layers that their dense versions on CPU and shows tremendous speedup for the last half of layers (it is even faster than GPU inference of dense layers Fig. \ref{fig:gpu_time}). The reasons for that fact should be further investigated, but one possible important detail is that the last layers inputs are small enough to fit into the processors top level cache completely, which empirically leads to no cache misses at all. If this is the case, the algorithm might be efficiently extended to the larger inputs by splitting them into groups - this approach should also be further investigated. After all, sparsity itself showed promising results in the model size reduction(\cite{sp_state}, \cite{on_the_fly}, \cite{lottery_ticket}, etc.) but lacked the efficient implementation, so the proposed algorithm might become a key to make these approaches applicable in practice.

The straightforward GPU implementation, however, doesn't show such impressive results (Fig. \ref{fig:gpu_speedup}). Anyway, it is already fast enough to obtain some speedup in practice and has the potential for improvement.
\par
Surprisingly, the most tough bound for the speedup for now is convolution with the low rank part of the decomposition. Even for quite high levels of compression the straightforward pytorch implementation gives no more than 4x speedup, also it is less efficient for small convolutions and large input (Fig. \ref{fig:lr_speedup_x1}) and does not give any speedup for GPU inference. This operation should be implemented more efficiently, especially for small layers to make the decomposed models more useful. Also it should be noted that this decomposition works better for larger input sizes (Fig. \ref{fig:lr_speedup_x1}, Fig. \ref{fig:lr_speedup_x2}, Fig. \ref{fig:lr_speedup_x3} for 214x214, 418x418 and 632x632 picture sizes passed through the network). Thus in some cases this approach may be used only for the last layers of the network, while the others are compressed using some other method.
\par
With current low rank and sparse convolution implementations the decomposition is less effective for smaller layers of the network: we pay with huge accuracy drop for minor speedup (\ref{fig:acc_drop}), while compression level stays almost the same. For now it may be reasonable to use this compression method for big layers only, while the other layers may be compressed using some other method. In future more efficient implementation may make this method effective for the whole network compression.

\section{Future work}
\subsection{Implementation efficiency}
As it was highlighted in Chapter \ref{discussion}, the efficient inference implementation for low rank part is needed. The raw version of sparse convolution should be implemented in the form that allows using it as a part of some deep learning framework (pytorch for example) and extended to use more that one core. Both of these operations should be implemented efficiently on GPU.
\par
The reasons for the tremendous speedup of sparse convolution for small input sizes should also be investigated. It is is a cache size issue, we may find out that there are some CPUs that can infer sparse networks efficiently. Also we may search for architectures that allow huge degree of sparsity and fit into these speedup conditions.

\subsection{Architecture choice}
The decomposition process described in Section \ref{compression_process} contains multiple heuristics related to the decomposition error choice and the decomposition order. These heuristics may be improved by investigating the relation between the decomposition parameters and the accuracy drop after fine tuning. Ideally this relation should let us search for the best parameters for each layer in terms of compression or speed up with given accuracy drop without the need to fine-tune on each iteration (which takes quite a long time with currently chosen heuristic search). Also, the sparsity rate through-over this work was fixed to 1\%, because this value already gave reasonable results, but it definitely should be better investigated in future.

\subsection{Decomposition improvement}
EPC was used to stabilize the CP decomposition terms and it gave some performance gain, but this is not the only way to make low rank part easier to fine tune. For example, authors of \cite{cp} report that NLS algorithm for CP decomposition gives much better results than ALS. The fine tuning strategies may also vary, for example, learning rates depending on matrix norms might be used. Also there are other types of decomposition applicable to the network compression task (for example, tucker decomposition \cite{tucker_compression}), that might be extended to low rank and sparse form.

\subsection{Layer importance metric}
As it was shown in Chapters \ref{compression_error_dependecy} and \ref{error_tolerance}, different layers require different relative number of parameters to describe their internal structure and affect the model performance differently. These may give us a clue to the layer importance metric - layers (or block of layers) that do not affect the model performance much might be replaced with smaller versions. Also, this may give some insights about model architecture search.

\subsection{Binarization approach}
The authors of \cite{bulat_matrix_2019-1} applied tensor decomposition to train binary network to 5\% better accuracy than the current SOTA in binarization, but it is still about 20\% worse than the original model. On the other hand, papers related to "lottery ticket hypothesis" \cite{lottery_ticket}, \cite{malach_proving_2020}, etc. showed that sparse subset of the weights preserves most of the model quality. Low rank and sparse approach may be sufficient to train binary low rank part wile keeping sparse part, which is still small enough, but preserves the model quality on reasonable level. Moreover, this sparse part may also be binarized - while having already the sparse pattern it is only needed to apply sufficient scaling.

\appendix
\newpage
\section{Tables}
\noindent

\begin{table}
\centering
\tiny
\begin{tabular}{ |c|c|c| } 
 \hline
 layer index & input shape, (channels\_in, X, Y) & kernel shape, (channels\_out, channels\_in, $K_x$, $K_y$) \\
 \hline\hline
 0 & (3, 224, 224) & (64, 3, 7, 7) \\
1 & (64, 56, 56) & (64, 64, 1, 1) \\
2 & (64, 56, 56) & (64, 64, 3, 3) \\
3 & (64, 56, 56) & (256, 64, 1, 1) \\
4 & (64, 56, 56) & (256, 64, 1, 1) \\
5 & (256, 56, 56) & (64, 256, 1, 1) \\
6 & (64, 56, 56) & (64, 64, 3, 3) \\
7 & (64, 56, 56) & (256, 64, 1, 1) \\
8 & (256, 56, 56) & (64, 256, 1, 1) \\
9 & (64, 56, 56) & (64, 64, 3, 3) \\
10 & (64, 56, 56) & (256, 64, 1, 1) \\
11 & (256, 56, 56) & (128, 256, 1, 1) \\
12 & (128, 56, 56) & (128, 128, 3, 3) \\
13 & (128, 28, 28) & (512, 128, 1, 1) \\
14 & (256, 56, 56) & (512, 256, 1, 1) \\
15 & (512, 28, 28) & (128, 512, 1, 1) \\
16 & (128, 28, 28) & (128, 128, 3, 3) \\
17 & (128, 28, 28) & (512, 128, 1, 1) \\
18 & (512, 28, 28) & (128, 512, 1, 1) \\
19 & (128, 28, 28) & (128, 128, 3, 3) \\
20 & (128, 28, 28) & (512, 128, 1, 1) \\
21 & (512, 28, 28) & (128, 512, 1, 1) \\
22 & (128, 28, 28) & (128, 128, 3, 3) \\
23 & (128, 28, 28) & (512, 128, 1, 1) \\
24 & (512, 28, 28) & (256, 512, 1, 1) \\
25 & (256, 28, 28) & (256, 256, 3, 3) \\
26 & (256, 14, 14) & (1024, 256, 1, 1) \\
27 & (512, 28, 28) & (1024, 512, 1, 1) \\
28 & (1024, 14, 14) & (256, 1024, 1, 1) \\
29 & (256, 14, 14) & (256, 256, 3, 3) \\
30 & (256, 14, 14) & (1024, 256, 1, 1) \\
31 & (1024, 14, 14) & (256, 1024, 1, 1) \\
32 & (256, 14, 14) & (256, 256, 3, 3) \\
33 & (256, 14, 14) & (1024, 256, 1, 1) \\
34 & (1024, 14, 14) & (256, 1024, 1, 1) \\
35 & (256, 14, 14) & (256, 256, 3, 3) \\
36 & (256, 14, 14) & (1024, 256, 1, 1) \\
37 & (1024, 14, 14) & (256, 1024, 1, 1) \\
38 & (256, 14, 14) & (256, 256, 3, 3) \\
39 & (256, 14, 14) & (1024, 256, 1, 1) \\
40 & (1024, 14, 14) & (256, 1024, 1, 1) \\
41 & (256, 14, 14) & (256, 256, 3, 3) \\
42 & (256, 14, 14) & (1024, 256, 1, 1) \\
43 & (1024, 14, 14) & (512, 1024, 1, 1) \\
44 & (512, 14, 14) & (512, 512, 3, 3) \\
45 & (512, 7, 7) & (2048, 512, 1, 1) \\
46 & (1024, 14, 14) & (2048, 1024, 1, 1) \\
47 & (2048, 7, 7) & (512, 2048, 1, 1) \\
48 & (512, 7, 7) & (512, 512, 3, 3) \\
49 & (512, 7, 7) & (2048, 512, 1, 1) \\
50 & (2048, 7, 7) & (512, 2048, 1, 1) \\
51 & (512, 7, 7) & (512, 512, 3, 3) \\
52 & (512, 7, 7) & (2048, 512, 1, 1) \\
 \hline
\end{tabular}
\caption{Resnet50 layers shapes}
\label{table:lshapes}
\end{table}

\begin{table}
\centering
\tiny
\begin{tabular}{ |c|c|c|c| } 
 \hline
 decomposition error bount & top1 accuracy & top5 accuracy & compression \\
 \hline\hline
 0.5 & 70 & 89.5 & 2.84 \\
0.6 & 65.2 & 86.7 & x3.6 \\
0.8 & 47 & 72.4 & x6 \\
 \hline

\end{tabular}
\caption{Decomposition with fixed error bound}
\label{table:fix_decomp}
\end{table}

\begin{table}
\centering
\begin{tabular}{ |c|c| } 
 \hline
 compression & speedup \\
 \hline\hline
1.5x & 1.07x \\
2x &  1.16x \\
3x & 1.34x \\
5x & 1.6x \\
10x & 2x \\
20x & 2.33x \\
 \hline
\end{tabular}
\caption{Decomposition with fixed error bound (x2 input size)}
\label{table:lr_speedup_x2}
\end{table}

\begin{table}
\centering
\begin{tabular}{ |c|c| } 
 \hline
 compression & speedup \\
 \hline\hline
1.5x & 1.09x \\
2x &  1.2x \\
3x & 1.4x \\
5x & 1.7x \\
10x & 2.11x \\
20x & 2.49x \\
 \hline
\end{tabular}
\caption{Decomposition with fixed error bound (x3 input size)}
\label{table:lr_speedup_x3}
\end{table}

\clearpage
\newpage

\clearpage
\section{Figures}

\begin{figure}
    \centering
    \includegraphics[width=0.4\textwidth]{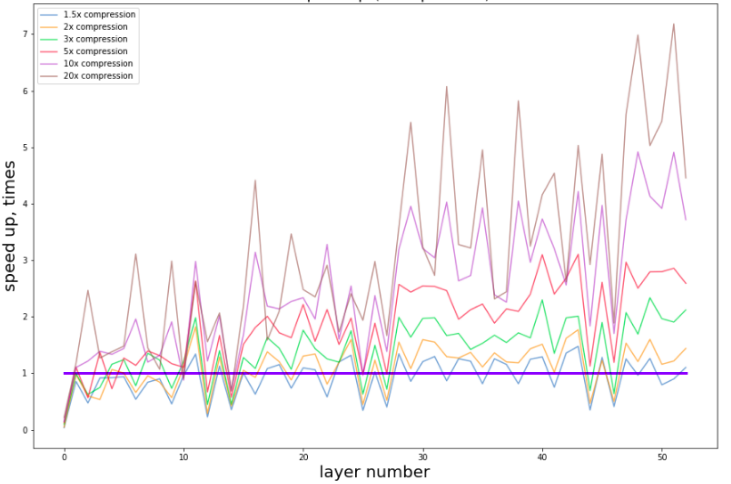}
    \caption{Resnet50 LR layers speedup on Intel Core i3-8130U CPU (x2 input size)}
    \label{fig:lr_speedup_x2}
  \hfill
\end{figure}

\begin{figure}
    \centering
    \includegraphics[width=0.4\textwidth]{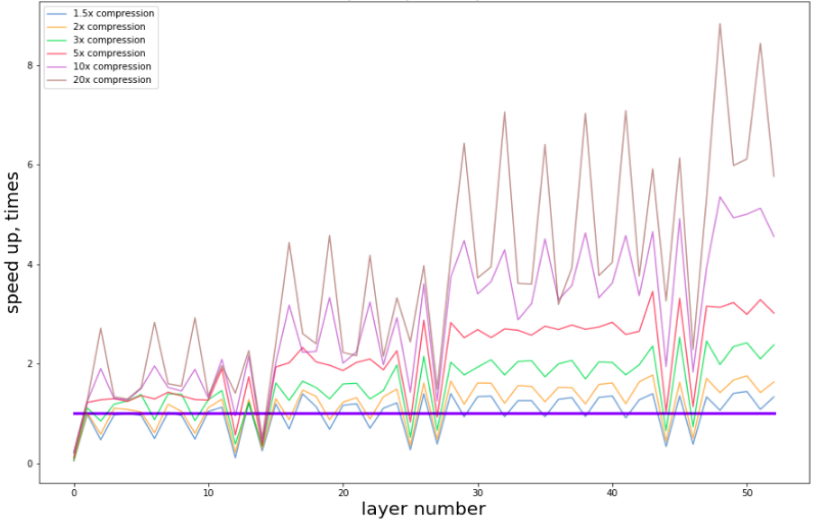}
    \caption{Resnet50 LR layers speedup on Intel Core i3-8130U CPU (x3 input size)}
    \label{fig:lr_speedup_x3}
\end{figure}

\begin{figure}
    \centering
    \includegraphics[width=0.4\textwidth]{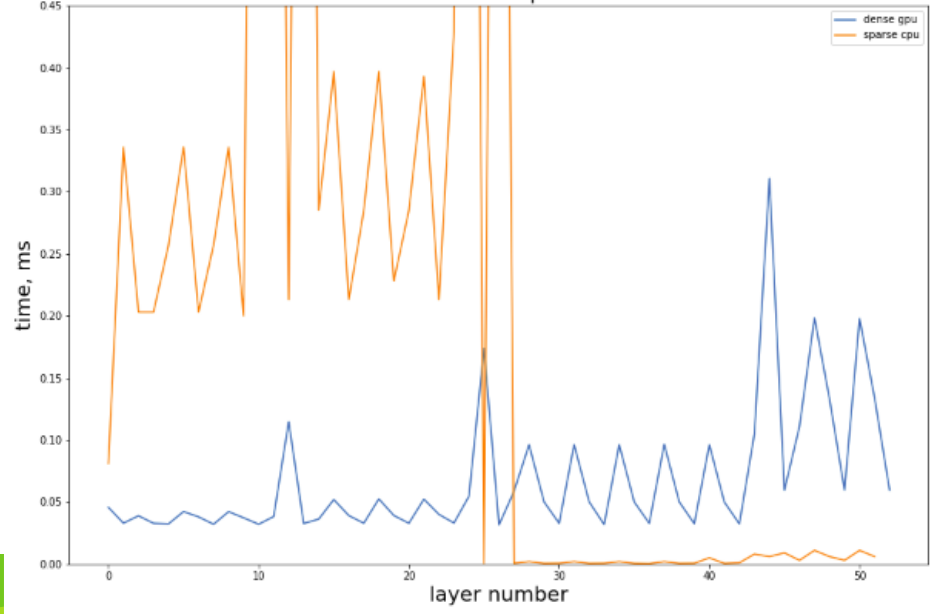}
    \caption{Sparse convolution Intel Core i3-8130U CPU inference time compared to dense convolution Nvidia 1080Ti GPU inference time}
    \label{fig:sp-gpu-time}
\end{figure}

\begin{figure}
        \centering
        \includegraphics[width=300pt]{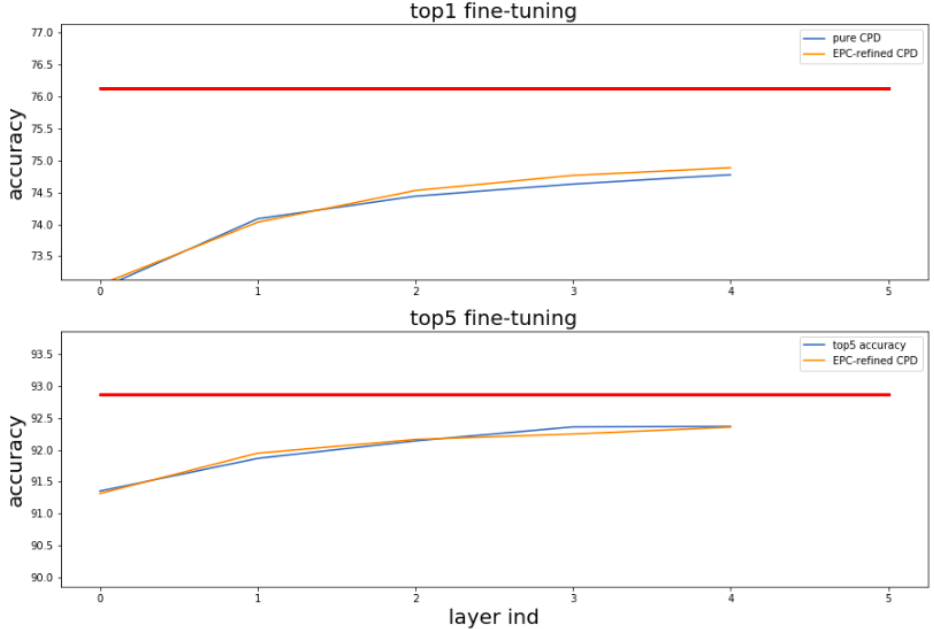}
        \caption{EPC fine tuning compared to raw CPD}
        \label{fig:epc}
\end{figure}

\bibliographystyle{jfm}
\bibliography{bib}


\end{document}